\documentclass{article}

\usepackage{PRIMEarxiv}

\usepackage[utf8]{inputenc} 
\usepackage[T1]{fontenc}    
\usepackage{hyperref}       
\usepackage{url}            
\usepackage{booktabs}       
\usepackage{amsfonts}       
\usepackage{nicefrac}       
\usepackage{microtype}      
\usepackage{lipsum}
\usepackage{fancyhdr}       
\usepackage{graphicx}       
\graphicspath{{media/}}     

\usepackage{amsmath,amssymb,amsfonts}
\usepackage{algorithmic}
\usepackage{graphicx}
\usepackage{textcomp}
\usepackage{subcaption}
\usepackage{bm}

\newcommand{\figref}{Figure~\ref}

\newcommand{\etal}{\textit{et al}.}
\newcommand{\ie}{\textit{i}.\textit{e}.}
\newcommand{\eg}{\textit{e}.\textit{g}.}

\title{Costraint-aware Policy for Compliant Manipulation}

\author{
  Daichi Saito\\
  Tokyo Institute of Technology \\
  \texttt{saito.d.ah@m.titech.ac.jp} \\
   \And
  Kazuhiro Sasabuchi \\
  Applied Robotics, Microsoft \\
  \And
  Naoki Wake \\
  Applied Robotics, Microsoft \\
  \And
  Atsushi Kanehira \\
  Applied Robotics, Microsoft \\
  \And
  Jun Takamatsu \\
  Applied Robotics, Microsoft \\
  \And
  Hideki Koike \\
  Tokyo Institute of Technology \\
  \And
  Katsushi Ikeuchi \\
  Applied Robotics, Microsoft \\
}

\begin{document}
\maketitle

\begin{abstract}
Robot manipulation in a physically-constrained environment requires compliant manipulation. Compliant manipulation is a manipulation skill to adjust hand motion based on the force imposed by the environment. Recently, reinforcement learning (RL) has been applied to solve household operations involving compliant manipulation. However, previous RL methods have primarily focused on designing a policy for a specific operation that limits their applicability and requires separate training for every new operation. We propose a constraint-aware policy that is applicable to various unseen manipulations by grouping several manipulations together based on the type of physical constraint involved. The type of physical constraint determines the characteristic of the imposed force direction; thus, a generalized policy is trained in the environment and reward designed on the basis of this characteristic. This paper focuses on two types of physical constraints: prismatic and revolute joints. Experiments demonstrated that the same policy could successfully execute various compliant-manipulation operations, both in the simulation and reality.
We believe this study is the first step toward realizing a generalized household-robot.
\end{abstract}

\keywords{Compliant manipulation; Reinforcement learning; Learning-from-Observation}

\section{Introduction}
Many household operations require manipulating an object under a physically constrained environment, such as opening drawers and doors.
A robotic system performing such household operations must guarantee not to damage the object or environment. Therefore, the robot needs to adjust its hand motion during the execution based on the force imposed by the environment, \ie, constraint force. This manipulation is called \textit{compliant manipulation}~\cite{mason1981compliance}. 
There are an unpredictable amount of manipulations in the household environment; thus, the generalized controller to such manipulations is expected to realize a household-robot.

This study investigates the generalization capability of a policy trained with a single environment and reward using reinforcement learning (RL) to various unseen manipulations.
Although RL-based approach~\cite{yahya2017collective, gu2017deep, Rajeswaran-RSS-18, urakami2019doorgym, sun2021door} is more robust to the uncertainty associated with recognition of object information, such as pose, articulation, and shape than classical controllers~\cite{karayiannidis2016adaptive}, this requires a manual design of the training environment and reward specific to each manipulation. 
Thus, it is not scalable to the number of target manipulations.
This issue is caused by the lack of the generalization of the policy to the unseen manipulations because this approach handles each manipulation independently.


Manipulations can be classified based on a physical constraint.
In the previous study~\cite{ikeuchi2021semantic}, a manipulation group is defined to have a common admissible/inadmissible direction, along which the object can/cannot move.
For example, several manipulations, such as drawer-opening, plate-sliding, and pole-pulling, belong to the same group because the object's admissible motion directions are constrained under a linear guide. 
If an object tries to move in the inadmissible direction, the constraint exerts the force on the object. 
Thus, we notice that the manipulations grouped based on the constraint also have a common characteristic of the constraint force.
Since compliant manipulation operations are executed leveraging the force, we design a single policy generalized to various unseen manipulations on the basis of the characteristic of the constraint force.

\begin{figure}[htb]
  \centering
  \vspace{3mm}
  \includegraphics[width=\columnwidth]{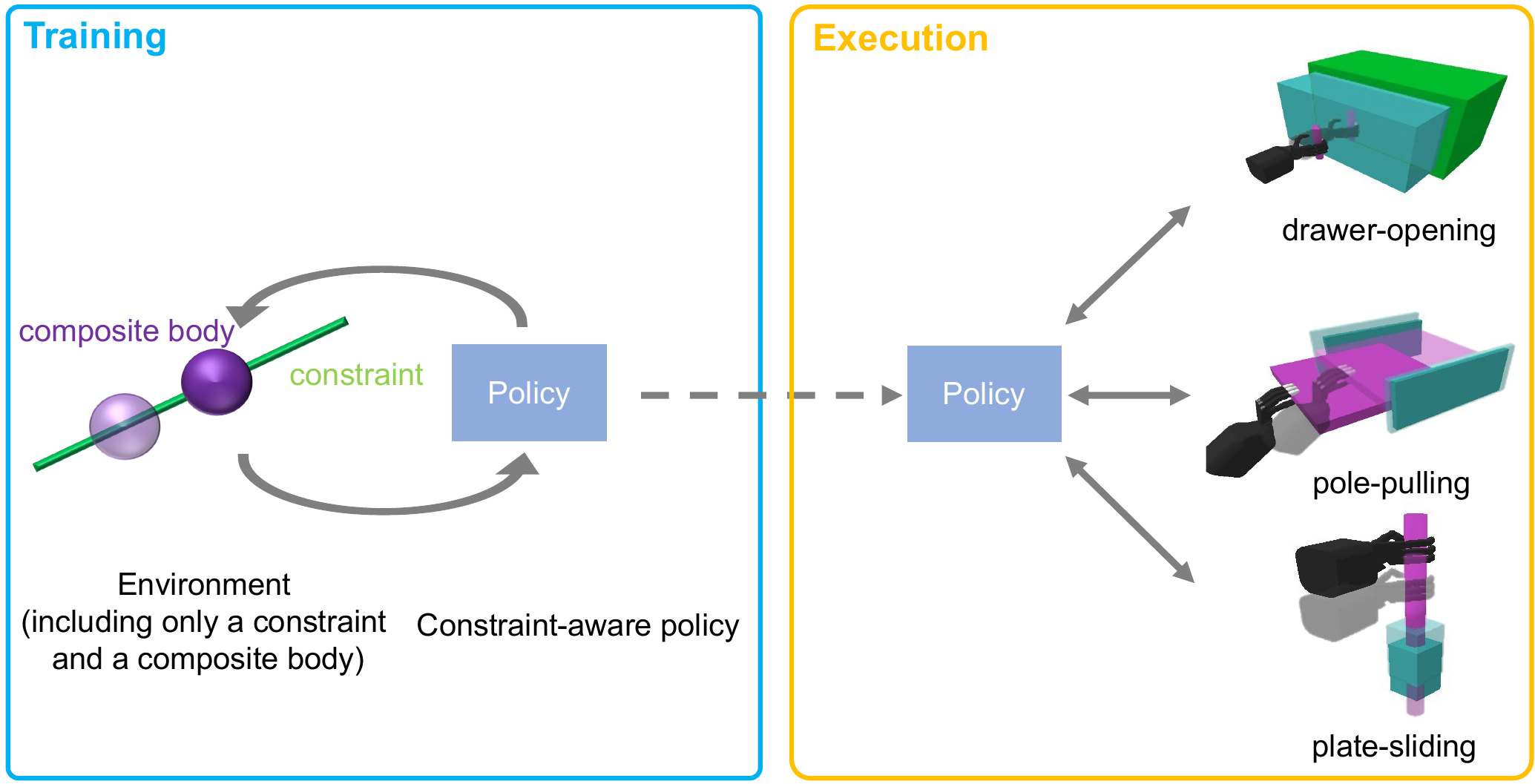}
  \caption{Concept behind constraint-aware policy. Various manipulations with a common physical constraint can be simplified as just a composite body and constraint, enabling the robot to obtain the constraint force, which we call the single-system condition. The constraint-aware policy is trained in the environment which consists of the body and its constraint. The policy can be applied to various manipulations with the same physical constraint under the single-system condition.}
  \label{fig:concept}
\end{figure}

We propose the \textit{constraint-aware policy} which estimates the object's admissible direction using the constraint force.
We train the policy to be generalized to unseen manipulations in the constraint group with a single environment and reward (right of \figref{fig:concept}).
These environment and reward are designed assuming \textit{single-system condition} (left of \figref{fig:concept}) that the robot hand and the object move in unison and can be regarded as the composite body, where the internal forces, such as frictional forces, are canceled out.
Thus, the policy can obtain the constraint force exerted on the object.
The environment is designed as a simplification of the real-world manipulations by extracting the common characteristic of the physical constraint critical to compliant manipulation operations, which is the key to the generalization.
The assumption is practically realistic because it can be easily satisfied by an execution design, such as moving the hand slowly.
Under the single-system condition, the estimation error of the admissible direction decreases in accordance with the reduction of the magnitude of the constraint force; thus, the reward is calculated only utilizing the magnitude.

In this study, we design the policy for the manipulation group with a prismatic and revolute joint which are representative constraints in the household environment. Under the constraint of a prismatic and revolute joint, the object has one-degree-of-freedom translation and rotation, respectively. 
In addition to the generalization within a group, we investigate the transferability to a different group.
Specifically, we consider transferring the policy for a prismatic joint to the manipulations with a revolute joint.
To reuse the policy, we discuss the common and uncommon aspect of a revolute joint compared to a prismatic joint.
\begin{itemize}
    \item The common aspect: circular motion can be considered as a series of infinitesimal linear motions.
    \item The uncommon aspect: the hand must rotate in conjunction with the object to achieve the single-system condition.
\end{itemize}
From the common aspect, we can apply the same constraint-aware policy so that the policy estimates the admissible direction in the both groups.
Whereas, owing to the uncommon aspect, the hand should rotate at execution only in manipulations with a revolute joint. To decide whether the hand needs to rotate or not, the type of the physical constraint, such as a prismatic or revolute joint, should be known.

To identify the constraint type, we leverage Learning-from-Observation (LfO)~\cite{ikeuchi1994toward, wake2021learning, wake2022interactive}. 
LfO provides the robot with hints for a manipulation through a multimodal one-shot human demonstration which includes a verbal instruction and hand movement.
The instruction contains semantic information that enables a robot to infer the constraint type of the manipulated object. For example, the verbal instruction of "open the refrigerator door" is associated with a revolute joint. 
At execution, the robot selects the policy corresponding to the obtained constraint type from the preliminary prepared policies.
In this study, we determine whether the physical constraint is a prismatic or revolute joint using the LfO system, and find out the necessity of the rotation of the hand.

We conducted experiments to investigate the generalization capability of trained constraint-aware policy to various unseen household-manipulations, such as a drawer-opening, plate-sliding, pole-pulling, door-opening, and handle-rotating, in the simulator.
We also compared the generalization with the classical controller~\cite{karayiannidis2016adaptive}, which is designed for the group with a prismatic joint.
As a result of this experiment, unlike the classical controller, the constraint-aware policy can be executed in various manipulations.
In addition, we evaluated the performance in the real-world using the policy and the LfO system, and demonstrated that the policy can be applied on a physical robot without additional training.

Toward a robot system capable of performing a wide range of manipulations, it is important to design the generalized policy for each manipulation group. 
Given that household manipulations can be classified based on their common constraints~\cite{ikeuchi2021semantic}, the key to the generalized policies is to design an environment and reward focusing on a common characteristic within each constraint group.
This study validated the concept of the constraint-aware policy for two fundamental physical constraints, those are a prismatic and revolute joint.
We believe this study is the first step of realizing the generalized household-robot.

The contributions of this study are as follows:
\begin{itemize}
    \item We proposed a constraint-aware policy which is trained using a single environment and reward and generalized to various unseen manipulations with a common physical constraint.
    \item We designed a simple training environment and reward function based on the constraint for the training of the constraint-aware policy.
    \item We demonstrated that unseen compliant manipulation operations can be executed on a physical robot using the constraint-aware policy and the LfO system.
\end{itemize}

The remainder of this paper is organized as follows. Section \ref{related} reviews related work and states the focus of this paper. Section \ref{method} introduces the constraint-aware policy. Section \ref{lfo} describes the details of LfO to apply the constraint-aware policy in practice. Section \ref{experiment} presents experiments for compliant manipulation using the constraint-aware policy in the simulation and reality. Section \ref{discussion} discusses the result of our experiment and an extensibility of our method to hardware-level reusability and other constraints. Section \ref{conclusion} concludes this paper.

\section{Related Work}\label{related}
In this study, we focus on a design of a policy which is robust to uncertainty associated with recognition, such as object pose and articulation, and object shape. In addition, we aim to train the policy which is generalized to various unseen manipulations with a single environment and reward using RL. The representative approaches of compliant manipulation are planning-based approach, classical closed-loop controller, and RL. In this section, we briefly review these approaches for compliant manipulation.

Previous research has focused on designing policies for opening drawers and doors. The pioneering work on door-opening is
\cite{nagatani1995door}, where robot motion is planned based on a known door model. In an unstructured environment, the model is unknown, and two methods can be used: geometry estimation and a closed-loop online controller to minimize force and torque. Several studies have been conducted on geometry estimation~\cite{klingbeil2010learning, pmlr-v100-abbatematteo20a, ruhr2012door, li2020category, liu2022articulation, jain2021screw}, where articulation pose is estimated from visual input, and a motion trajectory can be planned from this estimation result.
However, the estimation accuracy is insufficient for compliant manipulation (\eg, $\sim 20^\circ$ estimation error in a rotation-axis orientation on real-world data~\cite{jain2021screw}), and causes the planning-based approach to fail.
To deal with such estimation errors, 
other studies~\cite{kessens2010utilizing, jain2010pulling} have devised a robot mechanism for compliance.

Closed-loop controllers have been proposed in several studies, which can deal with uncertainty in geometry estimation~\cite{niemeyer1997door, schmid2008opening, chung2009door, karayiannidis2012adaptive, karayiannidis2016adaptive}. In \cite{niemeyer1997door}, an online controller was designed on the basis of a simple strategy in which the end-effector follows the path of the least force. Several studies have proposed online controllers based on this strategy~\cite{schmid2008opening, chung2009door, karayiannidis2012adaptive, karayiannidis2016adaptive}. These online controllers 
use the magnitude of force, which differs due to the change in the environment, and are not robust to the environmental change. 
To address this issue, we propose a constraint-aware policy using RL that can deal with uncertainty.
Classical controllers also have a problem that requires manual parameter tuning. Adaptive controller is the solution to tune the parameters for a specific manipulation~\cite{pilastro2016nonlinear, corra2017adaptive, pareek2023ar3n}. This adaptive tuning requires a real-world interaction between the robot and environment.
In the case of our study that the object is constrained to the environment, a large force is directly applied to the robot and the object under an estimation error. Thus, it is dangerous to determine the parameters through the real-world interaction, and the controller is not appropriate for this study.
Using the learning-based approach for compliant manipulation mitigates the issue on the parameter tuning.

Several studies have applied RL to train a policy for compliant manipulation~\cite{yahya2017collective, gu2017deep, Rajeswaran-RSS-18, urakami2019doorgym, sun2021door}. These studies focused on the design of policies by preparing the environment and reward for only a specific manipulation. For example, these studies prepare a door-opening environment and calculate an angle of the door as the reward.
For example, Urakami~\etal~proposed DoorGym, which is a training environment for generalizing the door-opening policy~\cite{urakami2019doorgym}. This trained policy can be generalized to doors with various doorknobs, lighting conditions, and environmental settings, but has focused on only a door-opening. Therefore, the trained policy is unable to be applied to other manipulations with the same constraint. 
There are several studies on RL which focus on designing a generalized policy for many varieties of manipulations~\cite{brohan2022rt, reed2022generalist}. However, this approach needs to take time and effort to prepare environments for all target manipulations to collect a large amount of data. In addition, this approach achieves an insufficient success rate on real-world application, and needs to fine-tune the policy for a specific manipulation.
In this study, we propose a policy generalized to manipulations with a common physical constraint, using a single environment and reward based on the common characteristic among these manipulations. 


\section{Method}\label{method}
In this study, we aim to train the policy generalized to various compliant manipulation operations which is required in many household manipulations. Toward this policy, we design a single environment and reward based on the common characteristic of the physical constraint within a manipulation group. In this section, we explain an approach to the learning of this constraint-aware policy.

This section is organized as follows.
Subsection \ref{grouping} explains the target manipulation group in this study. 
Subsection \ref{assumptions} states assumptions for executing the constraint-aware policy.
Subsection \ref{training} introduces the training method of the policy for the target manipulation group in Subsection \ref{grouping}.
Subsection \ref{sectionD} describes the technical details of satisfying the single-system condition which is one of the assumptions explained in Subsection \ref{assumptions}. These details are essential for an appropriate execution of the policy trained under the environment and reward in Subsection \ref{training}.

\subsection{Target manipulation group}\label{grouping}
In this study, we focus on manipulation groups with the physical constraints which are one-degree-of-freedom translation (prismatic joints) or rotation (revolute joints).
These physical constraints are representative in the household environment.
In the manipulations with a prismatic joint, such as drawer-opening, plate-sliding, and pole-pulling, the object's admissible motion directions are constrained under a linear guide.
As for the manipulations with a revolute joint including door-opening and handle-rotating, the admissible directions are constrained under a rotational axis.

Compliant manipulations of the same group have a common characteristic of the constraint force. A large force is exerted on an object when the object tries to move along the inadmissible direction. 
Since compliant manipulation operations can be achieved using the force, we achieve various unseen manipulations within the same group by a single policy based on such a characteristic of the force.


\subsection{Assumptions}\label{assumptions}
The constraint-aware policy in this study is executed on the following assumptions.

\textbf{Assumption 1:}
Single-system condition: The robot hand and object move in unison, where the internal forces between them are canceled out.

\textbf{Assumption 2:}
The inertial force on the manipulated object is negligible.

\textbf{Assumption 3:}
Friction in the joint mechanism is sufficiently weak such that the manipulated object can move smoothly along the desired trajectory.

\textbf{Assumption 4:}
The workplane of the robot hand and direction of the rotation axis are known; thus, the robot hand and manipulated object move on a known plane.

These assumptions can be fulfilled in the manipulations we are focusing on.
Assumptions 1 and 2 can be satisfied through the design of the manipulation, with Assumption 2 being satisfied by moving the manipulated object slowly.
Assumption 1 is satisfied by a grasp mechanism and an additional policy to decrease torque exerted on the object.
For more details of Assumption 1, see Section \ref{sectionD}.
Assumption 3 is satisfied by many household objects as they are designed for easy handling by humans. 
Finally, this study focuses on objects with only one prismatic or revolute joint, which are representative in household environments; thus, regarding Assumption 4, the workplane can easily be obtained. These can be obtained using Learning-from-Observation (LfO), where a human provides manipulation instructions to a robot through a one-shot demonstration~\cite{ikeuchi1994toward, wake2021learning}. We can calculate the workplane from human hand trajectories. For more details on Assumption 4, see Section \ref{lfo}.

\subsection{Training design under single-system condition}\label{training}
Deep RL is employed to design the control policy as it mitigates requirement of manual parameter tuning and is robust to uncertainties such as recognition error and sensor noise, unlike classical controllers~\cite{niemeyer1997door, schmid2008opening, karayiannidis2016adaptive}.

To design the control policy, we assume compliant manipulation as a Markov decision process and apply deep RL to train a constraint-aware policy. The Markov decision process has a state space $\mathcal{S}$, action space $\mathcal{A}$, state transition $\mathcal{T} : \mathcal{S} \times \mathcal{A} \rightarrow \mathcal{S}$, initial state distribution $\rho_0$, and reward $r : \mathcal{S} \times \mathcal{A} \rightarrow \mathbb{R}$. 
At each timestep $t$, an agent interacts with an environment with an action $a_t$ determined from state $s_{t}$, resulting in $s_{t+1}$ and $r_{t+1}$. 
The goal of RL is finding the optimal policy $\pi(a|s)$ that maximizes the cumulative reward $J(\pi) = \mathbb{E}_{\pi}[\sum_{t=0}^{T-1}\gamma^t r(s_t, a_t)]$, where $\gamma$ is the discount factor, $\gamma \in [0, 1)$, and $T$ is the episode length.

In this study, the robot hand moves along a motion direction $\bm d \in \mathbb{R}^3$ and observes a force $\bm F  \in \mathbb{R}^3$. We train the policy $\pi$ to estimate an optimal motion direction while the hand moves along the estimated direction.

\subsubsection{Training Environment}
\begin{figure}[htb]
  \centering
  \includegraphics[scale=1.25]{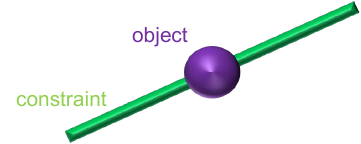}
  \caption{Training environment concept, consisting of the single composite body (purple sphere) and prismatic joint (green line).}
  \label{fig:env}
\end{figure}

The training environment is designed based on the single-system condition.
This environment consists of a single composite body and a prismatic joint (\figref{fig:env}). 
This composite body represents the robot hand and manipulated object under the single-system condition. 
At each timestep, a force exerted on the body $\bm F$ is obtained as a result of interaction between the body and constraint.
The constraint is represented as a constraint equation, and the force is calculated by solving the equation of motion which includes the constraint force~\cite{coumans2021}. 
The single-system condition guarantees that $\bm F$ measured at the robot wrist is identical to the constraint force on the body, as any internal forces between the hand and the object can be ignored.

This environmental design offers the advantage of a low simulation cost, as it is unnecessary to consider unstable factors, such as contact simulations between objects. 
This improves simulation speed and leads to faster training.
Furthermore, the policy trained in this environment can be easily adapted to different robot hands because it is independent of the specific characteristics of the robot hand itself.

\subsubsection{State and Action}
At timestep $t$, the state $\bm{s}_t \in \mathbb{R}^6$ consists of the normalized force obtained from a sensor
$\bar{\bm F}_t \in \mathbb{R}^3$ ($\bar{\bm F}_t = \frac{\bm F_{t}}{\lVert \bm F_{t} \rVert_2}$) and the motion direction of the robot hand $\bm d_t \in \mathbb{R}^3$. 
Utilizing the normalized force vector is important because the normalization makes the policy robust to a change in the magnitude of force, which is caused by an environmental change.
Note that if the constraint force is so small that they are negligible, various noises such as sensing errors and joint bending are amplified. In this study, we assume that the constraint force is constantly large enough to ignore these factors. In case that these factors are negligible, we should calculate a magnitude of the force fewer than a predefined threshold as zero value.
The action $\bm a_t \in \mathbb{R}^3$ is defined as an operation that modifies the direction of motion.
Given $\bm s_t$ and $\bm a_t$, the motion direction is updated using the following equation:
\begin{gather}
    \bm d_{t+1} = \frac{\bm d_t + \bm a_t}{\lVert \bm d_t + \bm a_t \rVert_2}
\end{gather}

\begin{figure}[htb]
  \centering
  \includegraphics[scale=0.5]{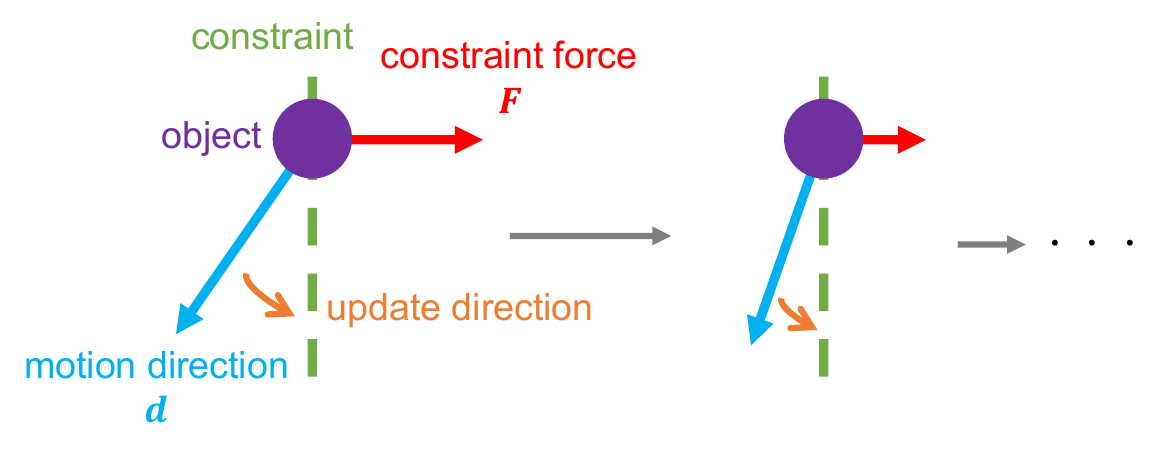}
  \caption{Updating motion direction using the constraint-aware policy. The purple circle and green line represent an object and its constraint, respectively. When the object tries to move in the inadmissible direction, the constraint force is exerted on the object. The motion direction is modified toward this force direction such that the force is reduced.}
  \label{fig:update}
\end{figure}

When the object tries to move in the inadmissible direction, the constraint force is exerted on the object. The policy should modify the motion direction toward this force direction such that the force is reduced.
As shown in \figref{fig:update}, the update of the motion direction by the optimal policy guarantees to adjust $\lVert \bm F \rVert_2$ resulting from the interaction between the object and the constraint.
Thus, the motion direction can be appropriately modified using the force direction. Note that the direction of the constraint force can be obtained under Assumption 3 where a friction in the joint mechanism is sufficiently weaker than the constraint force.

\subsubsection{Reward}
We train the constraint-aware policy to estimate the motion direction of the robot hand. To train the optimal policy, we should set an appropriate reward function based on the constraint. Thus, we consider the case that the motion direction is not along the constraint~(\figref{fig:update}).
In compliant manipulations with both the prismatic and revolute joints, if the robot hand does not move along the constraint, the constraint force is exerted by the physical constraint on the object. This force is minimized when the motion direction is along the constraint. 
Thus, we propose the reward $r_t$ represented by the constraint force $\lVert \bm F_t \rVert_2$:
\begin{gather}
    r_t = - \lVert \bm F_t \rVert_2
\end{gather}

\subsection{Technical details of satisfying the single-system condition when applying the policy to a robot}\label{sectionD}
The constraint-aware policy is trained and executed under the assumption of the single-system condition. To satisfy the single-system condition, the relative position and orientation between the robot hand and an object must be maintained. Two main challenges to satisfy this condition are identified: fingertip slipping, and lack of contact between the robot and object.

\subsubsection{Avoidance of fingertip slipping}
A violation of the single-system condition can occur if a large impulse force causes the robot's fingertips to slip on the manipulated object. This large impulse force is mainly caused in case that the robot hand tries to move in the inadmissible direction by the large amount of translation. Thus, to prevent the large impulse force, we implement the robot control system so that the robot hand moves slowly.
Moreover, fingertip slipping is likely to occur if the hand orientation remains constant during manipulation of a revolute joint where the orientation of the manipulated object changes. To avoid the slipping, we change the hand orientation based on the change in the motion direction, as follows. 
We define $q_t$ as the quaternion representing the hand orientation in the world coordinate system at time $t$; then, $q_{t+1}$ can be calculated using the following equation:
\begin{gather}
    q_{t+1} = \Delta q_t \otimes q_t
\end{gather}
where $\Delta q_t$ represents the quaternion rotating the angle between $\bm d_t$ and $\bm d_{t+1}$ around the outer product of $\bm d_t$ and $\bm d_{t+1}$.

\begin{figure}[htb]
  \centering
  \includegraphics[scale=0.75]{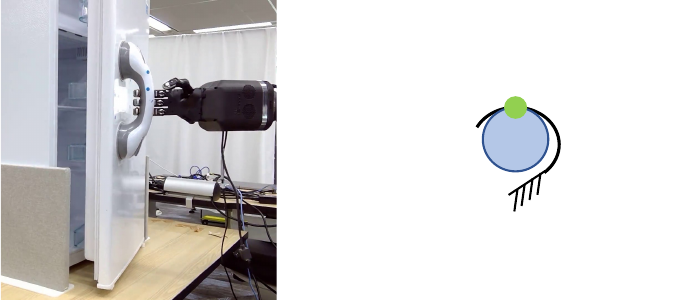}
  \caption{Door-opening with "Lazy-closure". A photograph of the actual manipulation is shown on the left. The right of the figure shows a diagrammatic representation of a robot grasping a handle with a Lazy-closure, where the blue and green circles indicate the handle and contact points, respectively, and the black arc is the gripper.}
  \label{fig:dooropening}
\end{figure}

This strategy does not necessarily guarantee to change the orientation of the hand completely in conjunction with the orientation of the object, and can be adopted only in case the relative orientation between the hand and manipulated object is not strictly fixed. An example case is door-opening with a lazy-closure, which is one of grasps~\cite{saito2022task}, as shown in \figref{fig:dooropening}. Using the lazy-closure, the contact regions remain constant and stable manipulation is ensured while opening the door, even though the relative orientation between the hand and manipulated object is not strictly fixed. However, when the relative orientation between the hand and manipulated object is strictly fixed, such as handle-rotating, a more precise method to change the hand orientation is required. Thus, we prepare an additional policy to maintain the single-system condition for this case. Further details are provided in the Appendix.

\subsubsection{Guarantee of hand-object contact}
The manipulated object and robot hand must be in contact throughout the manipulation to maintain the single-system condition. Contact is guaranteed if a non-zero constraint force is measured by a sensor on the wrist of the robot. Thus, the contact condition is ensured by applying a constraint force at the beginning of the manipulation and maintaining it throughout the manipulation. Specifically, the constraint force $\bm F$ fed into the policy is defined as the raw force value $\bm F_s$ offset by the force $\bm F_d$ (i.e., $\bm F = \bm F_s - \bm F_d$).

If the estimated displacement $\bm{d}$ is out of inadmissible directions between the robot hand and object, the hand goes away from the object. In such a case, the single-system condition is broken. In this study, we assume that the estimated displacement is always within the inadmissible directions between the robot hand and object.

\section{Learning-from-Observation System}\label{lfo}
Compliant manipulation is executed by combining our constraint-aware policy with the Learning-from-Observation (LfO), a system in which a human provides manipulation instructions to a robot through a one-shot demonstration~\cite{ikeuchi1994toward, wake2021learning}. 
In this study, the physical constraint, workplane and initial motion direction are obtained from a human demonstration for compliant manipulation. 
Using this system, we can satisfy Assumption 4, \ie, the workplane can be determined by leveraging the demonstration. This section describes the details of the LfO system applied in this study. 

\begin{figure}[t]
  \centering
  \includegraphics[width=\columnwidth]{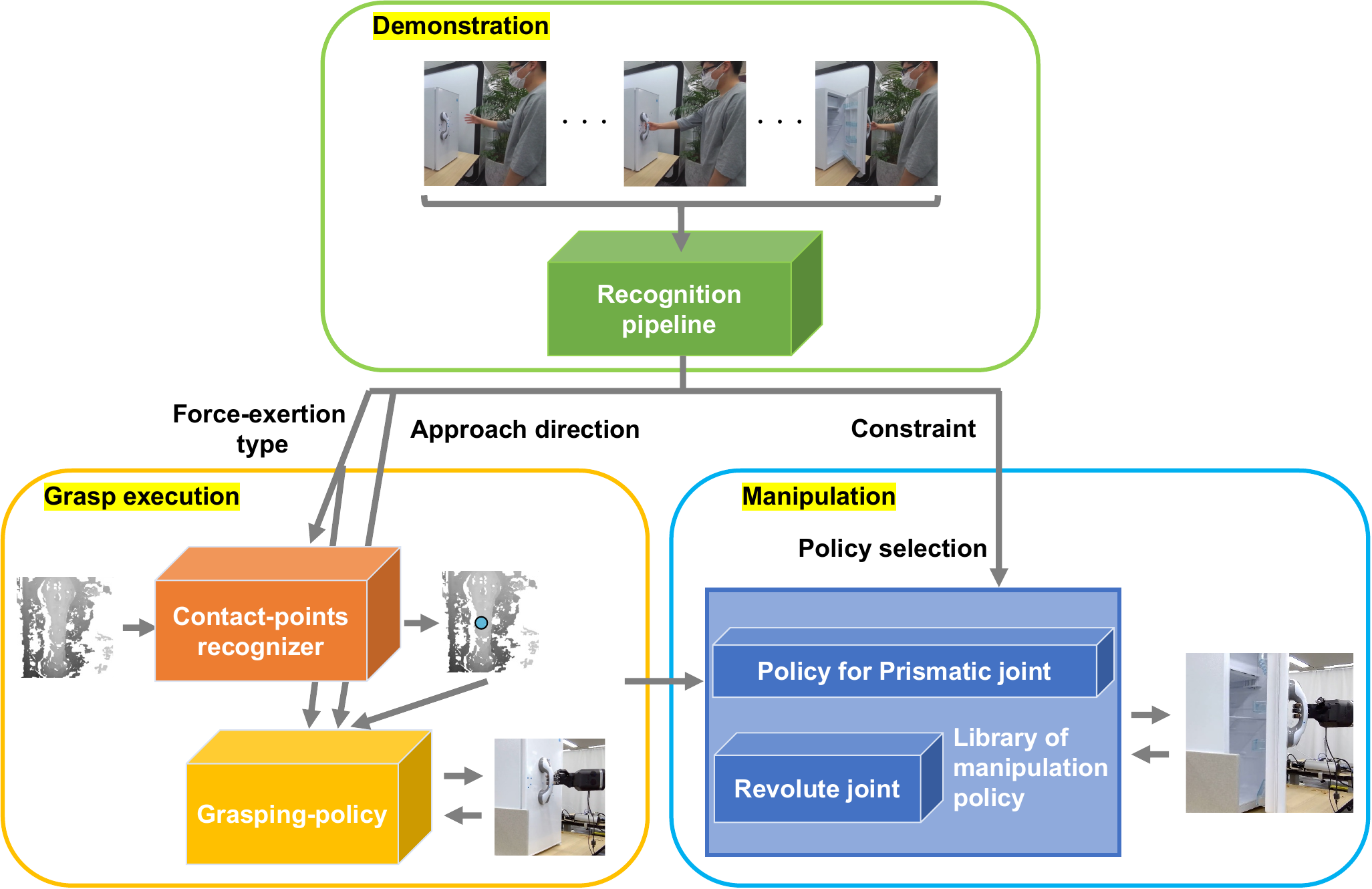}
  \caption{Flow of the LfO system combined with constraint-aware policy.}
  \label{fig:lfo}
\end{figure}

As shown in \figref{fig:lfo}, the LfO system consists of two phases: the demonstration phase and execution phase. The demonstration phase involves the LfO system obtaining a sequence of tasks from a human demonstration and assigning skill parameters to each task. During the execution phase, the system decodes the skill parameters into the execution commands.

In the demonstration phase, a human demonstration is encoded into a sequence of tasks using skill parameters~\cite{wake2022interactive}. The demonstration consists of an RGBD image sequence of a one-shot human demonstration and verbal instructions. In this study, the human demonstration is decomposed into several tasks, including the grasping and compliant manipulation within physical constraints (prismatic or revolute joint). The skill parameters of grasp and manipulation are also determined from the image sequence and verbal instructions. 

For grasping, the skill parameters include the force-exertion type and approach direction appropriate for the task situation~\cite{saito2022task}. A convolutional neural network (CNN)-based classifier recognizes one of the four force-exertion types based on the human hand image at the moment of grasp and the name of the object~\cite{wake2020grasp}. Similar hand shapes can be recognized as different force-exertion types using the name of the object. 

For compliant manipulation of a prismatic joint, the skill parameters include the workplane normal and initial motion direction. Meanwhile, the skill parameters of compliant manipulation for a revolute joint include the rotation radius, in addition to the workplane normal and initial motion direction. These parameters are calculated from changing the position of the human hand in the demonstration. The workplane normal and rotation radius are calculated using plane fitting and circular fitting, respectively.

In the execution phase, the robot executes the target task sequence by first grasping an object and then manipulating it. In the grasping, a contact-points recognizer and grasping policy are selected based on the force-exertion type obtained in the demonstration phase~\cite{saito2022task}. The recognizer and policy are previously trained for each force-exertion type. The contact-points recognizer has a simple CNN structure, where the input is the depth image of the target object and the output is the contact points to be grasped. The detected contact points are passed on to the grasping policy, and the grasp is executed. 

In the manipulation, a manipulation policy is executed. The manipulation policy is selected based on the constraints obtained in the demonstration phase. In the task involving the prismatic and revolute joints, the constraint-aware policy is applied. Note that, as described in Section \ref{sectionD}, in a task with a revolute joint, the hand orientation is changed to maintain the single-system condition because the orientation of the manipulated object changes during the manipulation. Therefore, the constraint type (prismatic or revolute) must be determined prior to manipulation.

\section{Experiment}\label{experiment}
We evaluated the performance of the proposed constraint-aware policy in the presence of errors in motion direction. We also confirmed the generalization capability of our policy for manipulations with a common constraint. In addition, we evaluated the feasibility of executing our policy and the LfO system on a physical robot. These evaluation processes are described in more detail below.

\subsection{Setup} \label{setup}
The training environment was implemented using PyBullet simulator~\cite{coumans2021} and the policy was trained using Microsoft Bonsai, a framework for RL\footnote{https://www.microsoft.com/en-us/ai/autonomous-systems-project-bonsai}.
The episode length of the training environment was set to five timesteps ($T = 5$). 
To simulate the uncertainty in the sensors, Gaussian noise was added to the observed force and motion direction at the first timestep. 
The proximal policy optimization (PPO) algorithm~\cite{schulman2017proximal} was used to train the policy. 
Batch size and learning rate were set to $6,000$ and $5\times 10^{-5}$, respectively. The policy $\pi_\theta$ is parameterized by a multilayer perceptron with two 256-dimensional hidden layers. A hyperbolic tangent ($tanh$) was used as the activation function as in~\cite{schulman2017proximal}.



The learned policy was tested using PyBullet simulator. The motion direction was updated every $100$ ms in the control loop, and the robot hand was moved by $1$ cm along the motion direction in each timestep. Each test started with the robot hand grasping the object, which was achieved using another RL policy~\cite{saito2022task}.

\begin{figure}[htb]
  \centering
  \includegraphics[scale=0.4]{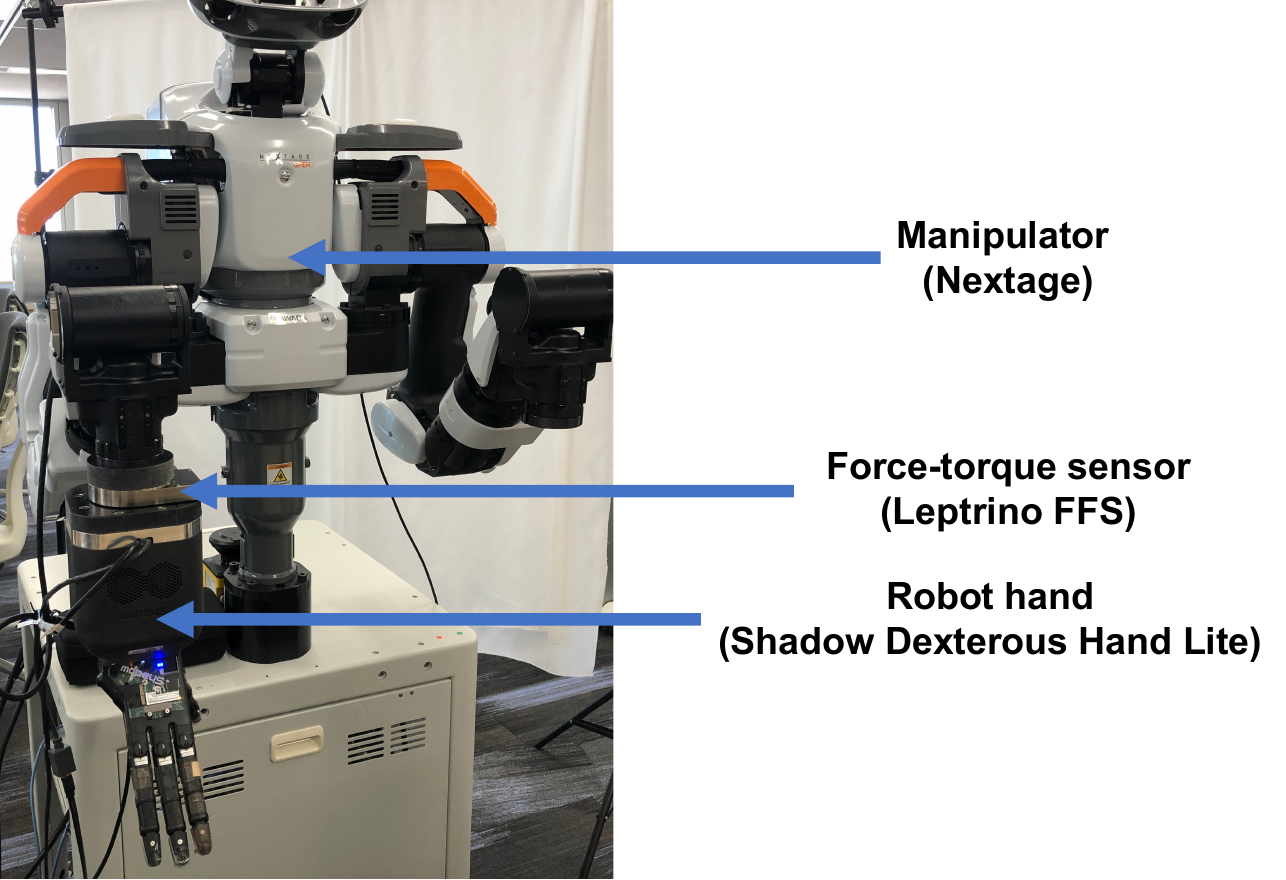}
  \caption{Robot setup, with force-torque sensor attached between the manipulator and robot hand.}
  \label{fig:robot}
\end{figure}

For the physical robot experiments, we utilized a Nextage\footnote{https://nextage.kawadarobot.co.jp/} robot with six degrees of freedom in its arms. A four-fingered robot hand, the Shadow Dexterous Hand Lite\footnote{https://www.shadowrobot.com/dexterous-hand-series/}, was attached to the robot. The Leptrino FFS series\footnote{https://www.leptrino.co.jp/product/6axis-force-sensor} was utilized as the force-torque sensor and attached between the manipulator and robot hand, as shown in \figref{fig:robot}. 

\subsection{Training in simulation}
\begin{figure}[htb]
  \centering
  \includegraphics[scale=0.5]{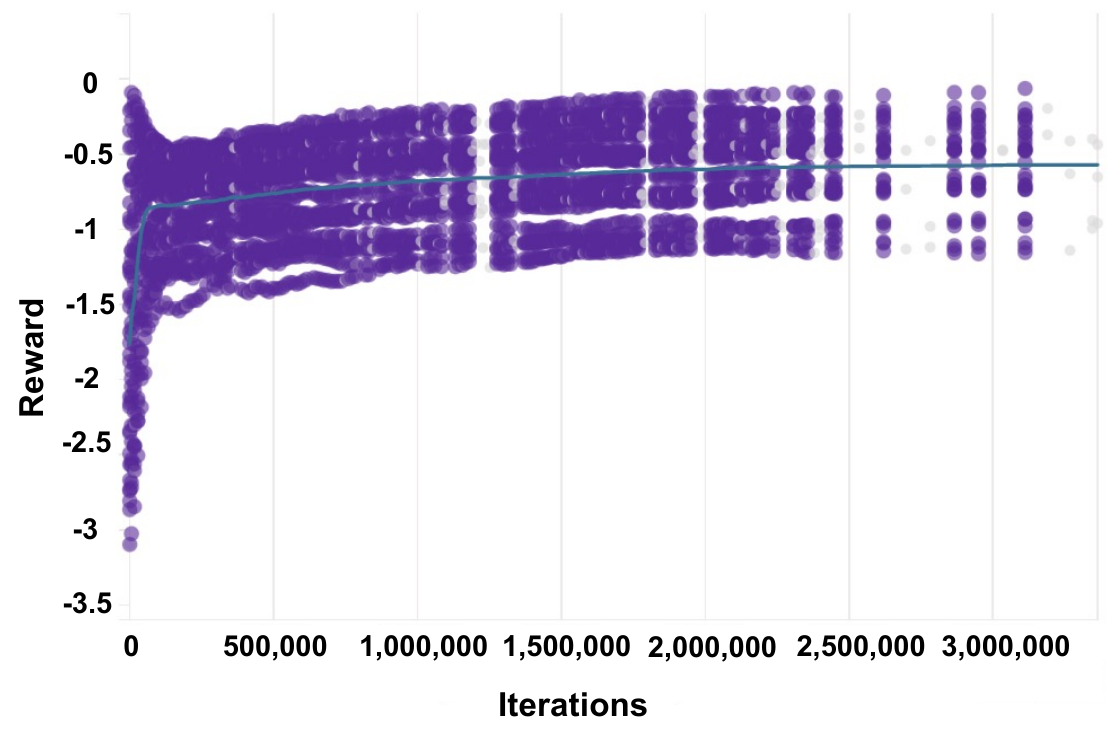}
  \caption{Learning curve of the constraint-aware policy. The blue line and dots are the mean reward of multiple episodes and reward of each episode, respectively. The purple dots represent the reward when the policy is saved.}
  \label{fig:lrcurve}
\end{figure}

The policy was trained in the simulation environment consisting of the object and prismatic joint.
The episode reward obtained by the RL agent increased as the training progressed, and the training was completed when the rewards converged (\figref{fig:lrcurve}).

\subsection{Policy performance in presence of motion direction errors}\label{experiment1}
A simulated drawer-opening environment was used to evaluate the performance of the proposed policy when the policy faced an error in the motion direction. The drawer was constrained by a prismatic joint, and the episode was considered completed when the drawer had been moved by $25$~cm. The handle of the drawer was grasped using a lazy-closure. 

\begin{figure}[htb]
  \centering
  \includegraphics[scale=0.4]{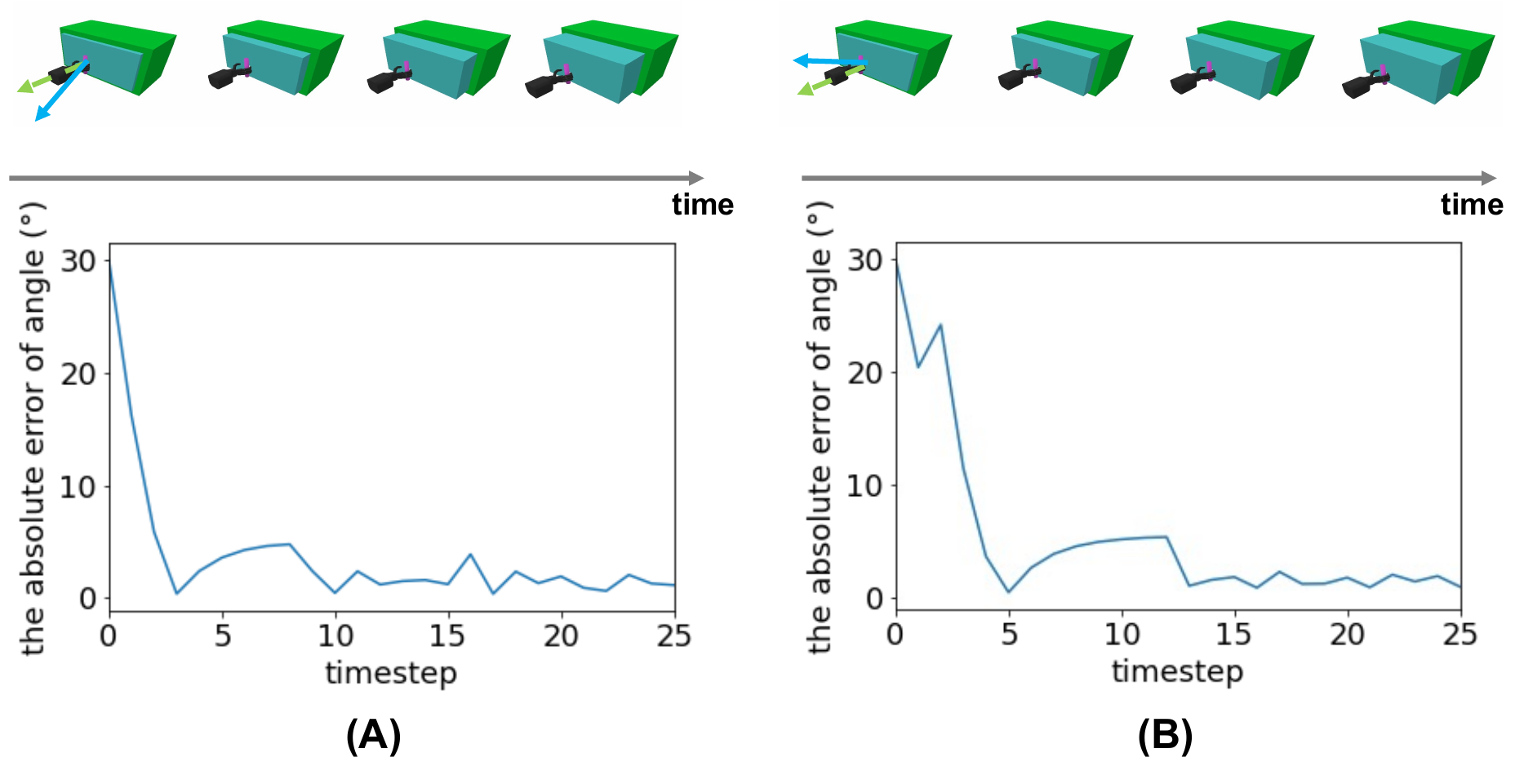}
  \caption{Policy performance in the presence of motion direction error. (A) The initial motion direction was set with a $30^{\circ}$ offset from the constraint direction. (B) Initial motion direction was set with a $-30^{\circ}$ offset from the constraint direction. The upper panel shows the resulting simulated drawer-opening. The lower panel shows the change in the relative angle between the admissible constraint direction (green arrow) and the current motion direction (blue arrow).}
  \label{fig:drawer}
\end{figure}

The results are presented in \figref{fig:drawer}, where the initial motion direction was set with a $30^{\circ}$ (\figref{fig:drawer}~(A)) or $-30^{\circ}$ (\figref{fig:drawer}~(B)) offset from the admissible constraint direction. 
In both cases, the drawer-opening was successfully executed. The curves represent the change in the relative angle between the admissible constraint direction and the current motion direction. The angles converged to near $0^{\circ}$. 
This result indicates that the proposed constraint-aware policy could estimate the motion direction from the direction of constraint force.

\subsection{Comparison of proposed and classical controller for various manipulations}
To evaluate the generalzaition capability of the proposed constraint-aware policy for various manipulations, we compared it with a state-of-the-art classical controller~\cite{karayiannidis2016adaptive}. Our constraint-aware policy and classical controller were executed on three manipulations with a prismatic joint: (A) drawer-opening, (B) plate-sliding, and (C) pole-pulling. These manipulations were selected because they require different force-exertion-types for grasp, such as active-force, passive-force, and lazy closure. 
These force-exertion-types cover the types that need no regrasping~\cite{saito2022task}. 
The initial motion direction was set with a $-30^{\circ}$ offset from the constraint direction, similar to the conditions reported in Section \ref{experiment1}. 

\begin{figure}[htb]
  \centering
  \includegraphics[scale=0.38]{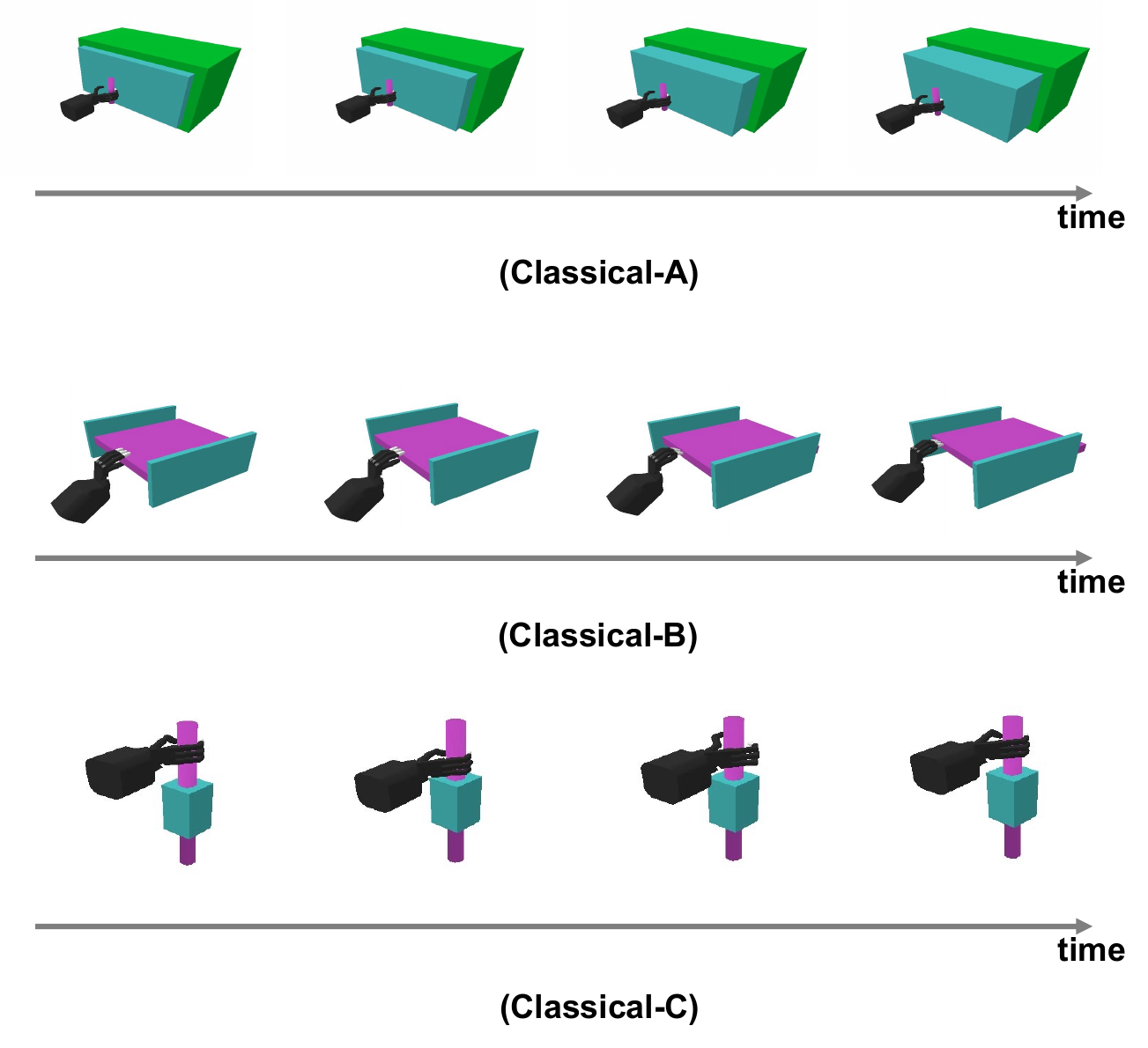}
  \caption{Execution of three manipulations using the classical controller~\cite{karayiannidis2016adaptive}. (A): drawer-opening, (B): plate-sliding, (C): pole-pulling.}
  \label{fig:comparison_classic}
\end{figure}

The results of the classical controller are shown in \figref{fig:comparison_classic} (Classical-A), (Classical-B) and (Classical-C). 
We manually tuned the control parameters for drawer-opening and used the same parameters for plate-sliding and pole-pulling. 
The controller succeeded in drawer-opening but not in plate-sliding or pole-pulling.
Since the motion direction could not be modified, the robot hand could not maintain its grasp, and consequently, the robot hand failed to manipulate the objects.
This is because the parameters were tuned for the magnitude of the sensed force, which differs according to the 
environment and the force-exertion-type.
The magnitude of the force varies depending not only on the grasp but also on the friction coefficient between the hand and object, object weight, damping coefficient of the finger joints, and sensor noise. Therefore, the magnitude of the force should not be used for robust execution.

\begin{figure}[htb]
  \centering
  \includegraphics[scale=0.38]{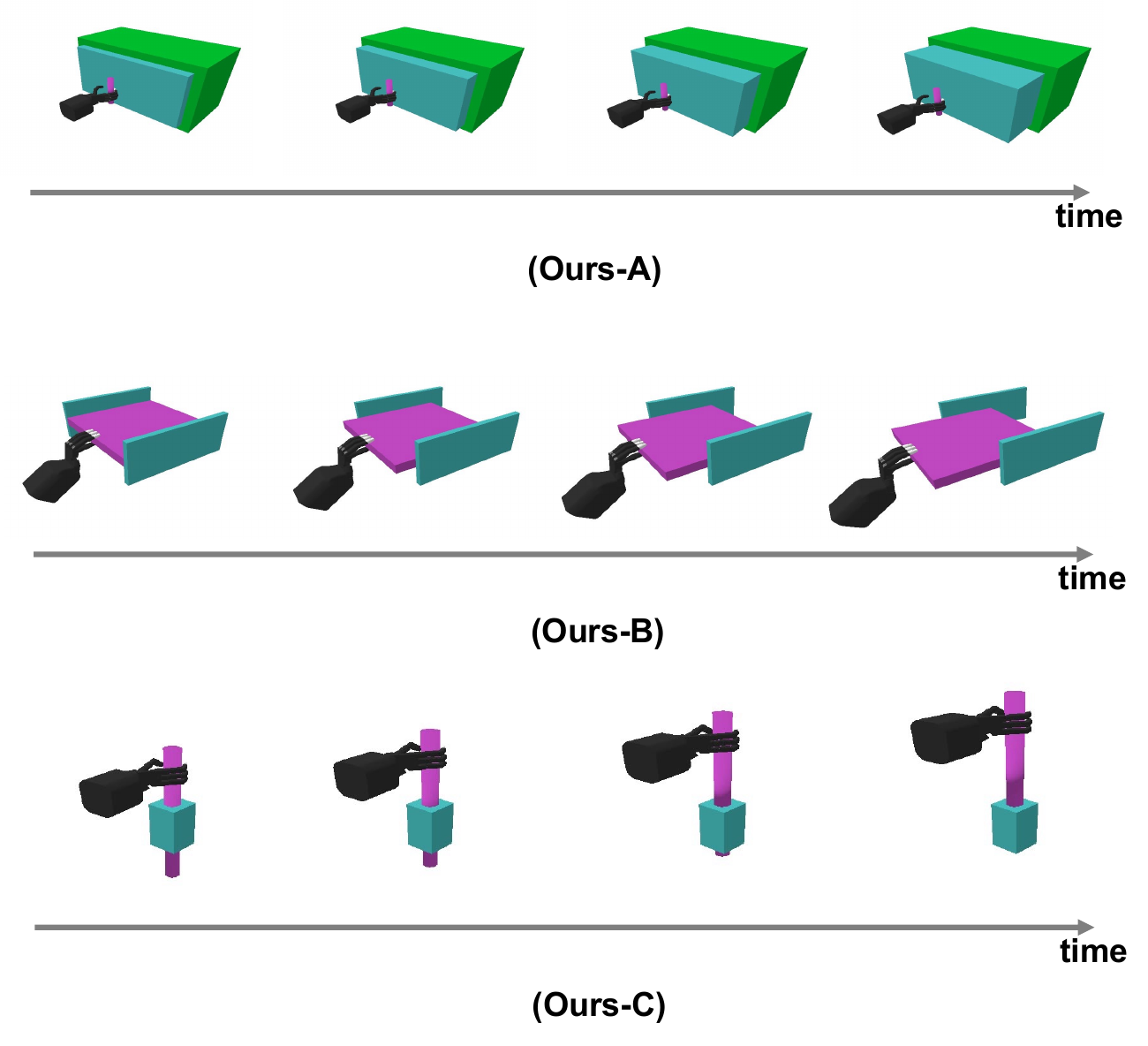}
  \caption{Execution of three manipulations using our constraint-aware policy. (A): drawer-opening, (B): plate-sliding, (C): pole-pulling.}
  \label{fig:comparison_ours}
\end{figure}

The results of our constraint-aware policy are shown in \figref{fig:comparison_ours} (Ours-A), (Ours-B), and (Ours-C). Unlike the classical controller, our constraint-aware policy succeeded in all three manipulations. This is because the utilized state includes the normalized force, instead of the raw force which is not robust to the change of the environment and force-exertion-type.
Using the normalized force makes the policy robust to changes in grasp. 

\subsection{Policy performance for manipulations with a revolute joint}
The proposed constraint-aware policy was executed in two different manipulations involving a revolute joint: door-opening and handle-rotating. The initial motion direction was set with a $15^{\circ}$ offset from the constraint direction.
Door-opening and handle-rotating were executed with a lazy and passive-force closure, respectively.

\begin{figure}[htb]
  \centering
  \includegraphics[scale=0.38]{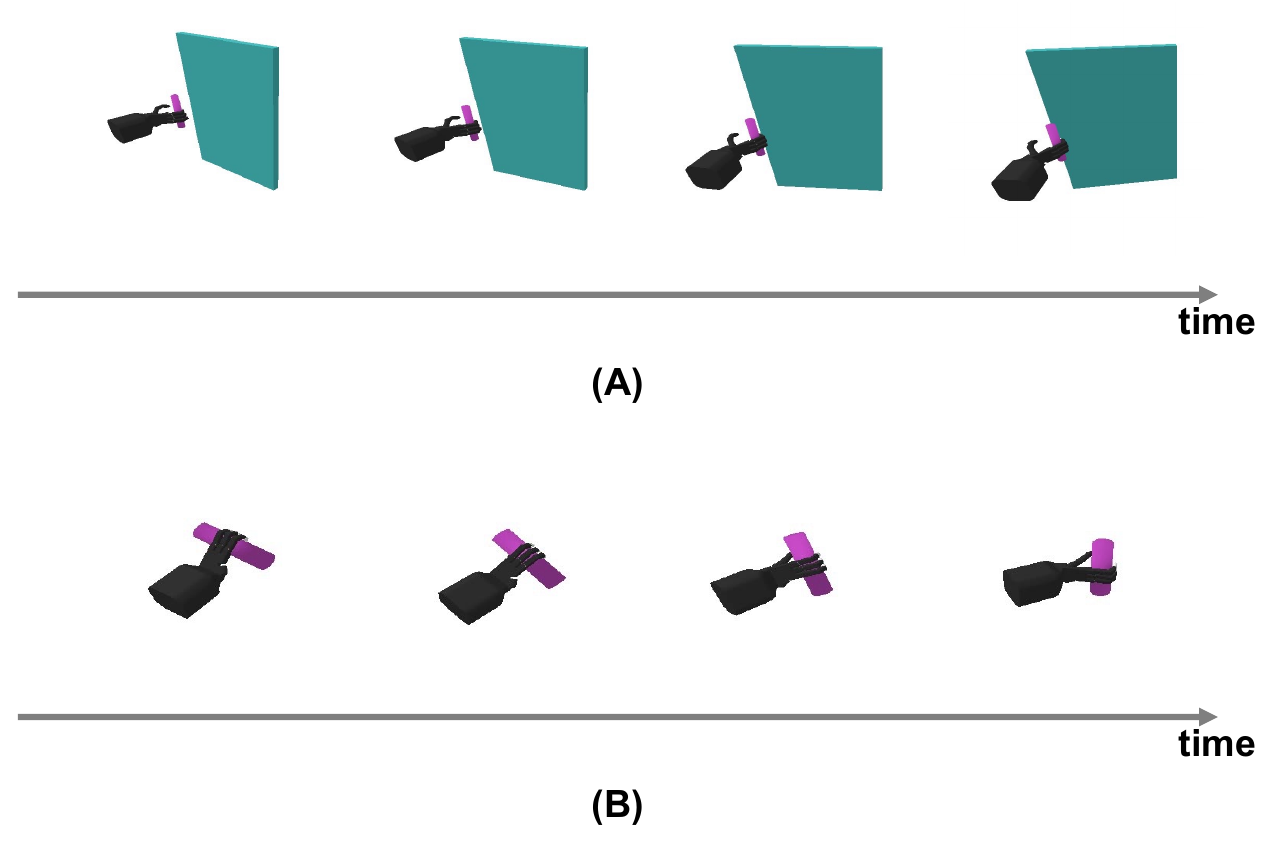}
  \caption{Execution of two manipulations with our constraint-aware policy. (A): door-opening, (B): handle-rotating.}
  \label{fig:revolute}
\end{figure}

The results are shown in \figref{fig:revolute}, demonstrating that the proposed policy could appropriately change the motion direction. Thus, our policy can be executed for manipulations with both prismatic and revolute joints under the single-system condition. In addition, the constraint force is directed from the handle to the rotation center, even when the rotation radius differs; thus our policy can be adopted for manipulations with varying rotation radius. 

\subsection{Compliant manipulation on a physical robot}
As mentioned prior, we combined our constraint-aware policy with the LfO system and executed it on a physical robot. In this method, the constraint was recognized from verbal instructions. It is important to identify the constraint because this is utilized to determine whether the hand rotates in conjunction with the manipulated object. In addition, the workplane and initial motion direction were determined. Grasp-Manipulation-Release sequence could be executed by incorporating constraint-aware policy into such an LfO system in the real-world.

\begin{figure}[tb]
  \centering
  \includegraphics[scale=0.4]{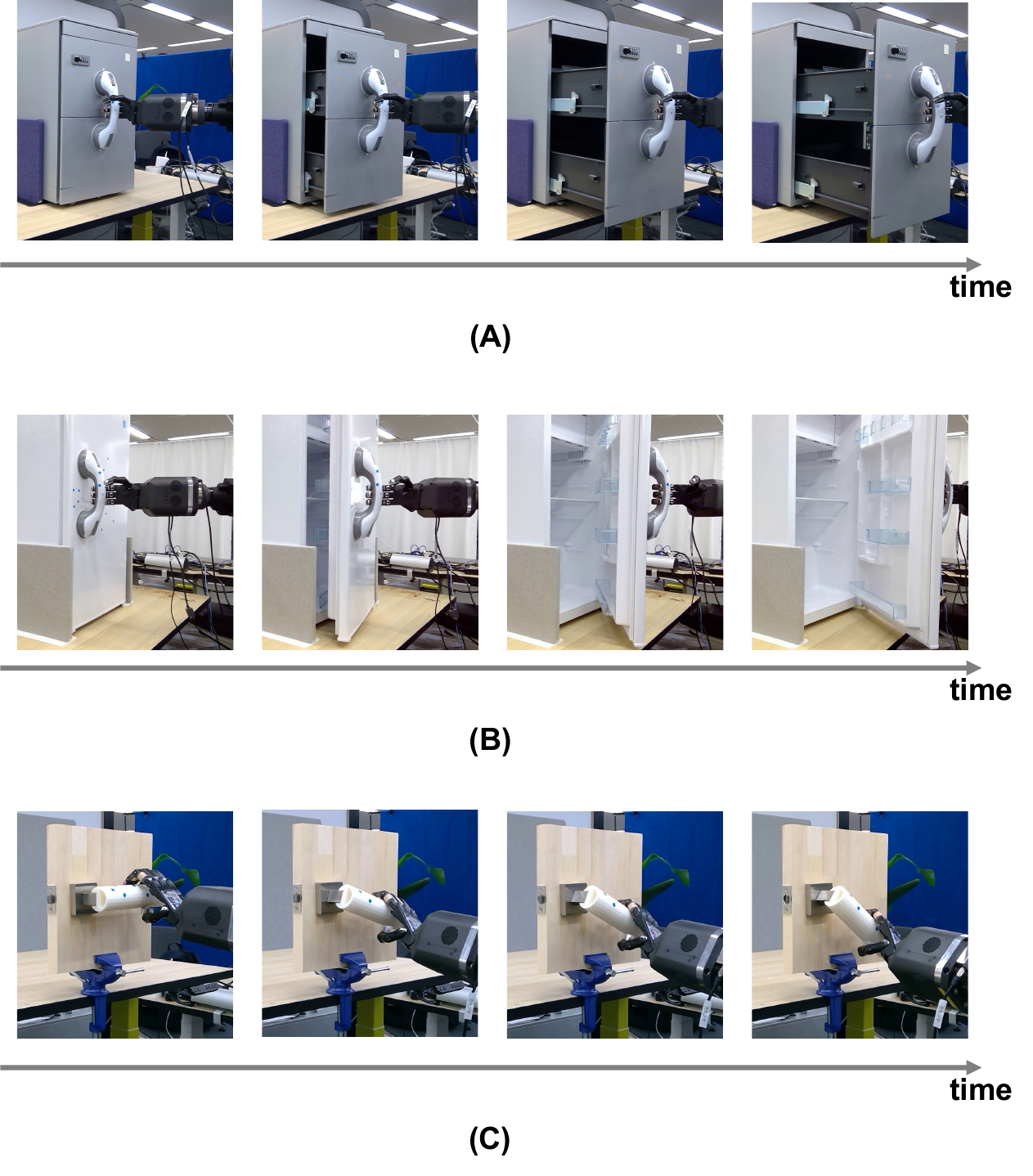}
  \caption{Applying the proposed constraint-aware policy for three manipulations using a physical robot: (A) drawer-opening, (B) door-opening, (C) handle-rotating.}
  \label{fig:execution}
\end{figure}

\figref{fig:execution} shows the successful execution of the three manipulations: (A) drawer-opening, (B) door-opening and (C) handle-rotating, in the real-world using our constraint-aware policy and the LfO system. Our policy uses a normalized force rather than a raw force, 
thereby reducing the gap between simulation and reality.
Consequently, our policy can be applied to the real-world without additional training. 

\begin{figure}[htb]
  \centering
  \includegraphics[scale=0.4]{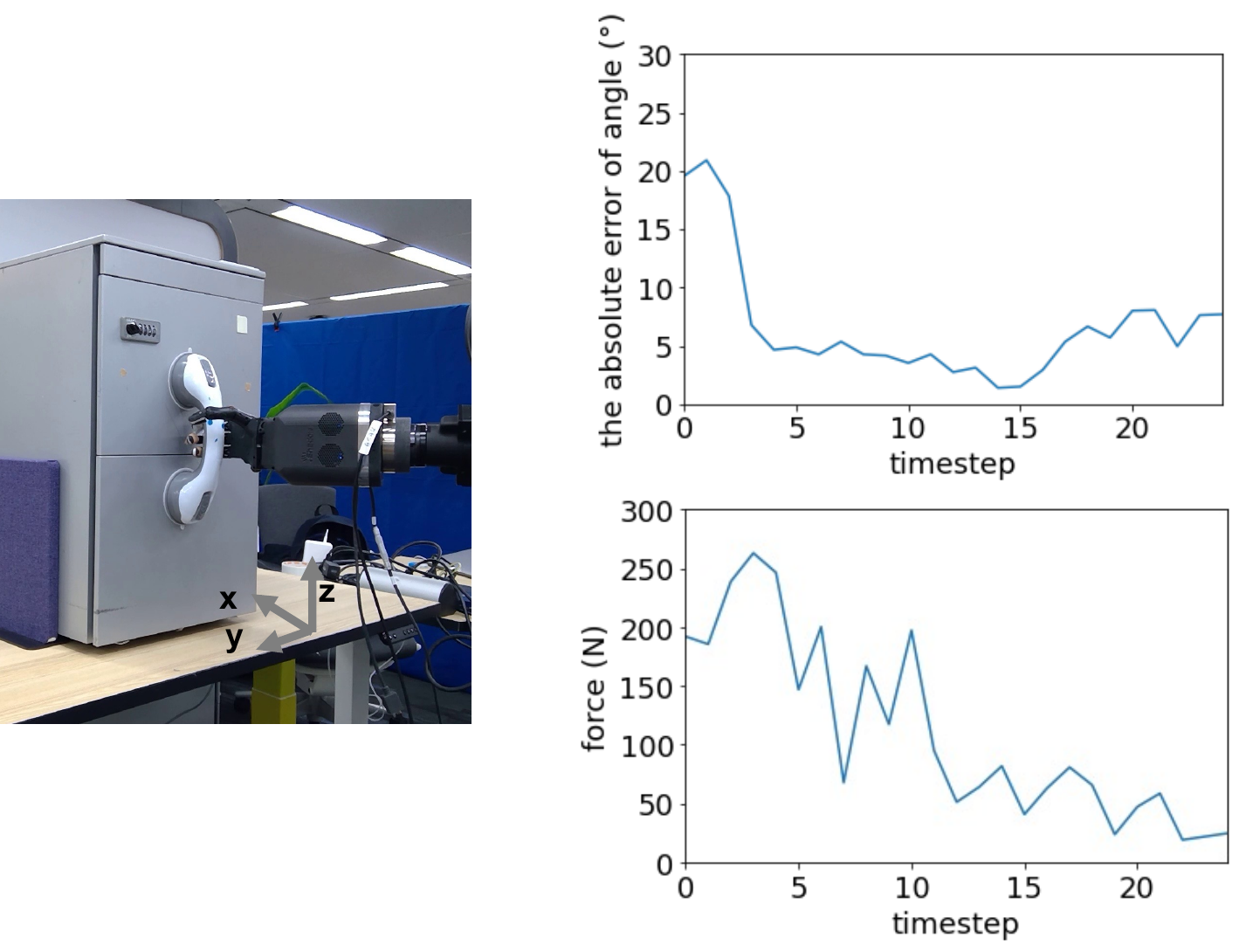}
  \caption{Execution of drawer-opening using proposed constraint-aware policy. Upper left: coordinate system. Upper right: change in relative angle between the estimated motion direction and admissible motion direction $(-1, 0, 0)$. :Lower right: the change in the magnitude of force recorded by the wrist force-torque sensor.}
  \label{fig:drawerforce}
\end{figure}

The left side of \figref{fig:drawerforce} shows the coordinate system used during the manipulation, while the upper-right side illustrates the change in the relative angle between the admissible motion direction $(-1, 0, 0)$ and the estimated motion direction. The lower-right chart in \figref{fig:drawerforce} illustrates the change in the magnitude of the force obtained by the wrist force-torque sensor. These results indicate that the angle and the magnitude of the force were being reduced during execution of the drawer-opening.

\begin{figure}[htb]
  \centering
  \includegraphics[scale=0.4]{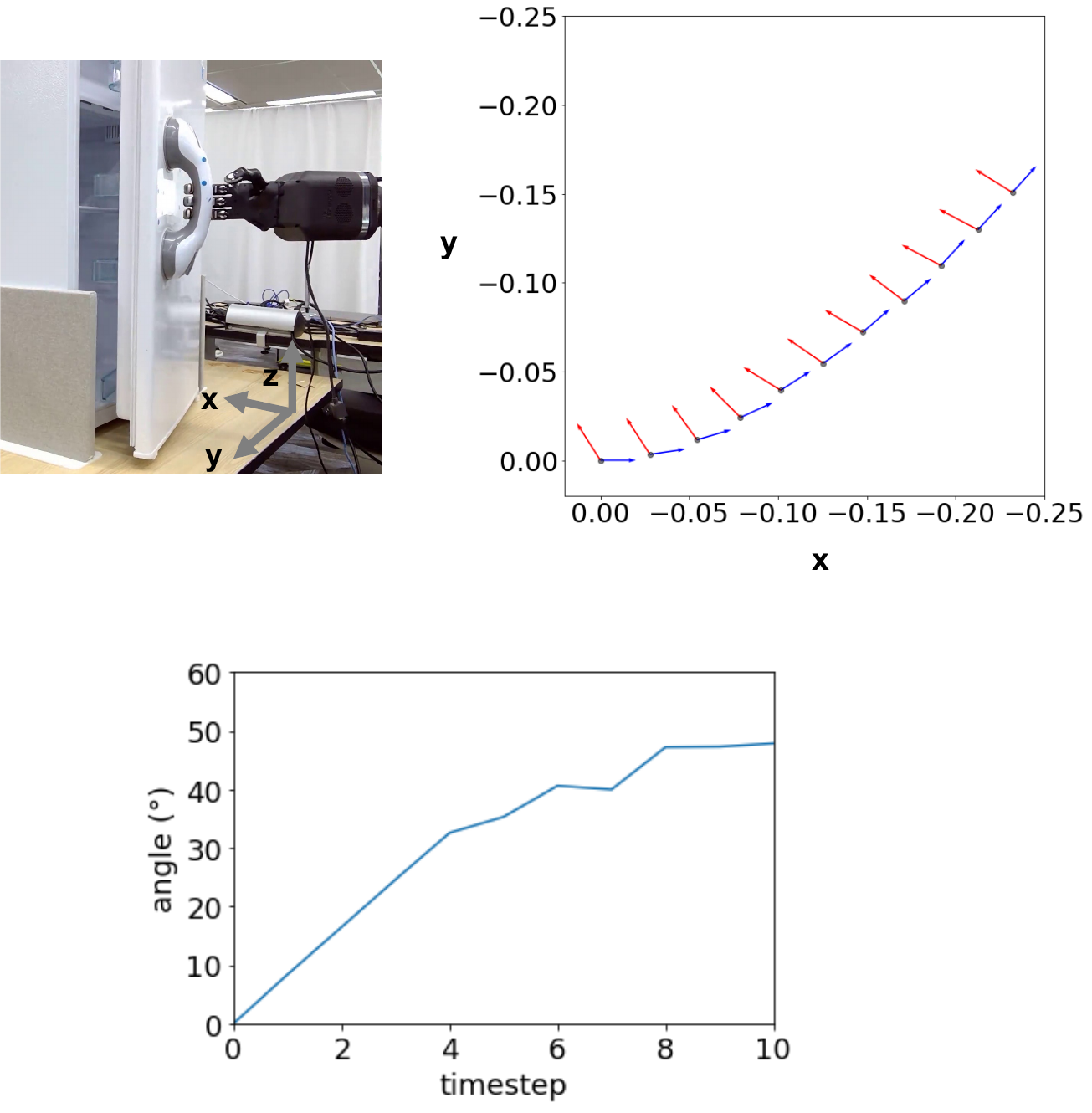}
  \caption{Execution of door-opening using proposed constraint-aware policy. Upper-left: coordinate system with origin on the index fingertip. Upper right: transition of the index fingertip position (black circle), motion direction (blue arrow) and force direction (red arrow) in meters. Lower: relative angle between initial motion direction $(-1, 0, 0)$ and the motion direction.}
  \label{fig:fridge}
\end{figure}

In \figref{fig:fridge}, the upper-right chart illustrates the transition of the index fingertip position, motion direction, and force direction during the door-opening, while the upper-left panel shows the coordinate system used during execution, where the origin was the fingertip position of the robot's index finger at the beginning of the manipulation. The lower part of \figref{fig:fridge} shows the relative angle between the motion direction and initial motion direction $(-1, 0, 0)$. It is evident that the motion direction changed based on the observed force direction, resulting in the successful execution of the door-opening, as shown in the upper-right panel of \figref{fig:fridge}. It was observed that the angle between the initial and actual motion directions gradually increased from the lower of \figref{fig:fridge}, as expected. 

\section{Discussion}\label{discussion}
\subsection{Summary of the experiments}
We propose the constraint-aware policy that is trained using the direction of the constraint force exerted on the object and generalized to various unseen manipulations. 
In this experiment, we investigated an effectiveness of our policy for manipulations with a prismatic and revolute joint. The results revealed that our policy succeeded in the execution of various manipulations: drawer-opening, plate-sliding, poll-pulling, whereas the classical controller~\cite{karayiannidis2016adaptive} failed. 
Although the environment and reward are simple, the policy is generalized. 
In addition, our policy succeeded in door-opening and handle-rotating. Finally, our policy could be executed on a physical robot without additional training. These results suggest that our policy is generalized to manipulations with a prismatic and revolute joint.
There is a possibility that our policy can be applied to more manipulations with these joints.

In terms of parameter-tuning cost, our method has an advantage compared to the classical controllers~\cite{schmid2008opening, karayiannidis2016adaptive}. The classical controllers require manual tuning of control parameters, whereas parameters of our policy can be tuned through the training.
In terms of training cost, our method needs a single environment compared to other policies trained by reinforcement learning~\cite{urakami2019doorgym, brohan2022rt, reed2022generalist}. These methods prepare environments of target manipulation for the training (in this study, the number of environments are five), whereas our policy can be trained under the one simple single-environment which includes only a constraint and composite body. This is a benefit of the policy design based on the common characteristic of the constraint force within a manipulation group.

\subsection{Limitation}
\subsubsection{Violation of single-system condition}
The proposed constraint-aware policy could be implemented under the single-system condition. 
One violation example of the single-system condition is slipping between the fingertip and manipulated object. 
This slipping can occur owing to the large estimation noise of the motion direction, rotation axis, and rotation radius. These issues cannot be addressed by our policy alone. A possible solution is to design an additional policy for maintaining contact positions by utilizing dexterous finger motions that depend on the force exerted on the fingertip. To implement this additional policy, tactile sensors are required. This will be a subject of future research.

\subsubsection{Normalized force}
We assume that the inertial force and friction in the joint mechanism is weaker than the constraint force and negligible. In our experiment, this assumption was satisfied and our method could be adopted. However, there is a case that this assumption is not met. For example, the case is that the estimation error of motion direction is near $0^{\circ}$. In this case, the weak inertial force and friction are amplified when normalizing them. This causes a system instability. To avoid the instability, we should calculate a magnitude of the force fewer than a predefined threshold as zero values. The way to define threshold will be a subject of future research.

\subsection{Future directions}
\subsubsection{Hardware-level reusability}
In this study, we designed a constraint-aware policy that can be applied to robot hands without considering hardware specifications assuming the single-system condition. In contrast to conventional strategies, our policy was designed to be both manipulation-agnostic and hardware-independent. When using new hardware, robot programmers typically must modify software, which can be time-consuming. To address this issue, some software programs enabling reusability have already been developed~\cite{quigley2009ros, takamatsu2022learning}. The work in this study represents another contribution to this field: using our constraint-aware policy, hardware-level reusability can be achieved. To demonstrate reusability, future studies will validate the hardware-level reusability of the proposed policy.

\subsubsection{Constraint-aware policy for other constraints}
Many manipulations in a household environment can be grouped based on constraints~\cite{ikeuchi2021semantic}. This taxonomy includes manipulation groups with prismatic and revolute joints as well as those with other constraints. One solution for achieving various manipulations in a household environment is to design a policy for each manipulation group. For achieving various household manipulations, our concept of constraint-aware policy can be applied to other constraints.
It should be effective to consider various manipulations with the same constraint as one manipulation group and design a policy with an awareness of the constraint.

\section{Conclusion}\label{conclusion}
In this study, we proposed a constraint-aware policy that can be applied to various manipulations with a prismatic and revolute joint. We designed a training environment and a reward function to train the policy based on these constraints. The experimental results showed that the single policy could be executed on three manipulations with a prismatic joint (drawer-opening, plate-sliding, and pole-pulling), even when an estimation error in the motion direction was applied in the simulation. Unlike the classical controller, our policy achieved the robust execution against environmental changes. In addition, we could execute our policy on two manipulations with a revolute joint (door-opening and handle-rotating). Furthermore, three manipulations, drawer-opening, door-opening, and handle-rotating, were successfully executed on an actual robot without additional training. 

Although our policy was trained in the simple environment, our policy could be executed successfully on different manipulations. Previous reinforcement learning (RL) methods specially designed the environment and reward for each target manipulation, whereas our policy was widely applicable to various assumed situations. Thus, we successfully designed a policy generalized to manipulations constrained by a prismatic and revolute joint based on the constraint force which is a common characteristic between such manipulations.

Toward a robot system capable of executing a wide range of manipulations, it is crucial to design a generalized policy for each manipulation group. 
Household manipulations can be categorized according to their physical constraints~\cite{ikeuchi2021semantic}. 
The key to the generalized policies is to design an environment and reward focusing on a common characteristic within each group. 
This study validated the concept of a constraint-aware policy for both prismatic and revolute joints, which are fundamental in considering physical constraints. 
We believe this study is the first step toward realizing the generalized household-robot.



\appendix
\section{Additional policy}
\begin{figure}[htb]
  \centering
  \includegraphics[scale=0.4]{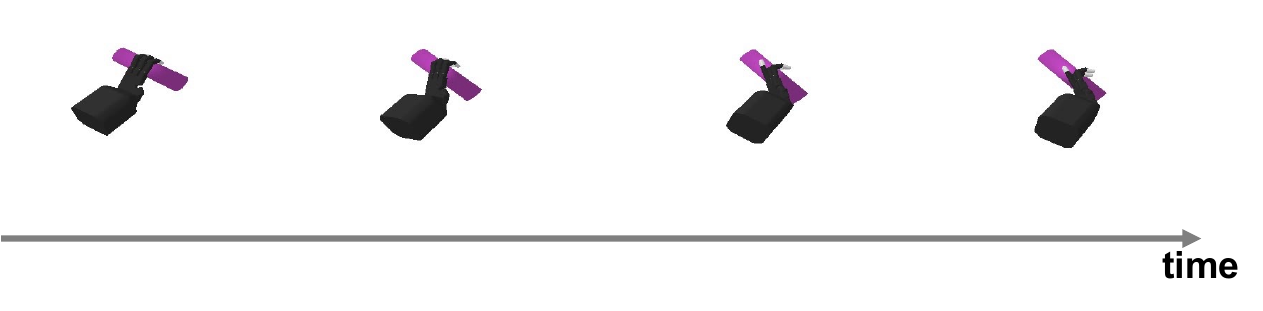}
  \caption{The failure result using the constraint-aware policy.}
  \label{fig:failure}
\end{figure}

We observed that the constraint-aware policy alone was unable to conduct handle-rotating with passive-force closure (\figref{fig:failure}). This failure occurred because it was impossible to maintain the single-system condition.
As described in the main text, when there is a change in the relative orientation between the robot hand and the manipulated object, torque is generated, causing slippage. 
The relative orientation between the hand and the manipulated object is strictly fixed, so the torque was generated due to a change in the relative orientation.
Thus, an additional policy was required that would enable the hand to rotate in conjunction with the manipulated object.

\begin{figure}[htb]
  \centering
  \includegraphics[scale=0.4]{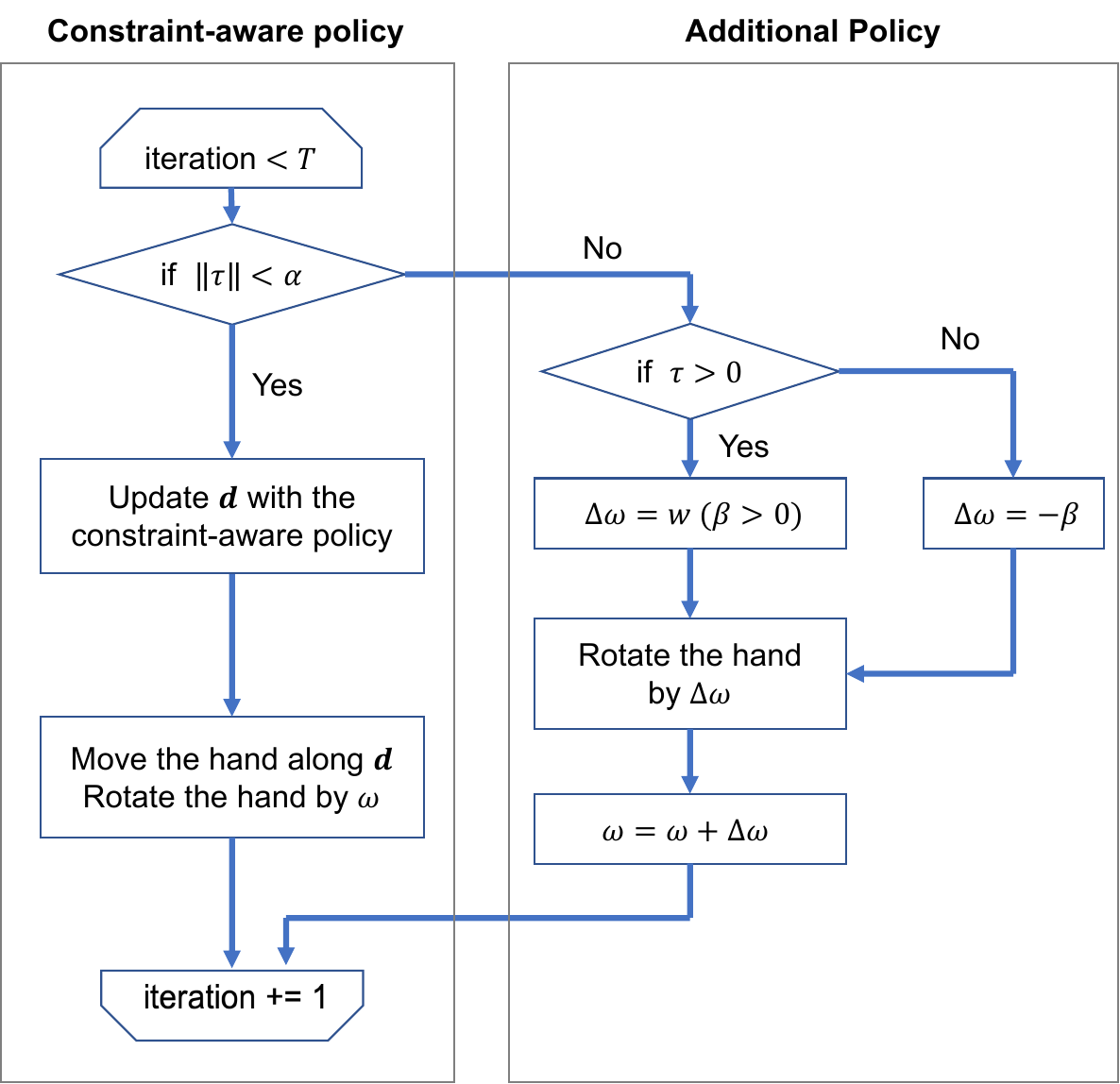}
  \caption{Combined policy for the case in which the hand cannot rotate freely around the rotation axis. The generalized policy is executed if the torque around the rotation axis $\lVert \tau \rVert$ is smaller than the threshold $\beta$. Otherwise, an additional policy is executed. $T$ is the episode length.}
  \label{fig:combined}
\end{figure}

An additional policy was developed to appropriately rotate the hand around
the center of the contact points to the handle. The rotation axis corresponds to the normal of the workplane.
This additional policy estimated the suitable amount of rotation $ w $ at each time step by the following process using the torque around the rotation axis $ \tau $.
\begin{enumerate}
    \item Rotate the hand by the current estimation of $ w $
    \item Decide the adjustment $ \Delta w $ as follows ($\beta > 0$): 
    \begin{displaymath}
        \left\{
        \begin{array}{ll}        
        \Delta w = 0  & (\lVert \tau \rVert \leq \alpha) \\
        \Delta w = \beta & (\tau > \alpha) \\
        \Delta w = -\beta & (\tau < \alpha)
        \end{array}
        \right.
    \end{displaymath}
    \item Update $w$ to $w+\Delta w$
\end{enumerate}
The initial value of $w$ is calculated using $w = \frac{v}{r}$, where $r$ is the rotation radius obtained from human demonstration and $v$ is the amount of translation in each timestep. 
The policy can calculate the excess or deficiency between $w$ and the suitable amount of rotation for one-step translation. The constraint-aware and additional policies are combined to execute the handle-rotating (\figref{fig:combined}). If $\tau$ is greater than $\alpha$ after the robot hand is translated and rotated simultaneously, the additional policy is implemented until $\tau$ is smaller than $\alpha$ to minimize forces other than the constraint force. Otherwise, the constraint-aware policy is implemented solely.

\figref{fig:revolute} (B) shows the successful result of handle-rotating using the combined policy. In the experiment, we set $\alpha=10, \beta=1^\circ$. 
The single-system condition was maintained by rotating the hand appropriately based on the torque. This result demonstrates that our constraint-aware policy, combined with the additional policy, can successfully execute a handle-rotating while maintaining the single-system condition. Although the additional policy requires manual tuning of the parameters to minimize the torque, these parameters have a higher interpretability for tuning than the control parameters of the classical controller.

\bibliographystyle{unsrt}  
\bibliography{templateArxiv}

\end{document}